\documentclass{article}

\usepackage{times}
\usepackage{graphicx} 
\usepackage{subfigure} 

\usepackage{natbib}

\usepackage{algorithm}
\usepackage{algorithmic}

\usepackage{epsfig}
\usepackage{graphicx}
\usepackage{amsmath}
\usepackage{amssymb}
\usepackage{tabularx}
\usepackage{booktabs}
\usepackage{amsthm}
\usepackage{color}
\usepackage{enumitem}
\usepackage{lipsum}
\usepackage{gensymb}

\usepackage{hyperref}

\usepackage[accepted]{icml2017_new}
\definecolor{olivegreen}{rgb}{0.33, 0.42, 0.18}

\usepackage[dvipsnames]{xcolor}

\icmltitlerunning{Graph-based Isometry Invariant Representation Learning}

\begin{document} 

\twocolumn[

\icmltitle{Graph-based Isometry Invariant Representation Learning}



\icmlsetsymbol{equal}{*}

\begin{icmlauthorlist}
%
\icmlauthorold{Renata Khasanova}{renata.khasanova@epfl.ch}
\flushleft{\icmladdressold{Ecole Polytechnique F\'ed\'erale de Lausanne (EPFL), Lausanne, Switzerland}}
\icmlauthorold{Pascal Frossard}{pascal.frossard@epfl.ch}
\flushleft{\icmladdressold{Ecole Polytechnique F\'ed\'erale de Lausanne (EPFL), Lausanne, Switzerland}}
\end{icmlauthorlist}


\icmlcorrespondingauthor{Renata Khasanova}{renata.khasanova@epfl.ch}

\icmlkeywords{deep learning, graph signal processing, machine learning, transformation invariant features}

\vskip 0.3in
]



\printAffiliationsAndNotice{}  


\begin{abstract} 
Learning transformation invariant representations of visual data is an important problem in computer vision. Deep convolutional networks have demonstrated remarkable results for image and video classification tasks. However,  they have achieved only limited success in the classification of images that undergo geometric transformations. In this work we present a novel Transformation Invariant Graph-based Network (TIGraNet), which learns graph-based features
that are inherently invariant to isometric transformations such as rotation and translation of input images. In particular, images are represented as signals on graphs, which permits to replace classical convolution and pooling layers in deep networks with graph spectral convolution and dynamic graph pooling layers that together contribute to invariance to isometric transformations. Our experiments show high performance on rotated and translated images from the test set compared to classical architectures that are very sensitive to transformations in the data. The inherent invariance properties of our framework provide key advantages, such as increased resiliency to data variability and sustained performance with limited training sets.
\end{abstract} 

\section{Introduction}


\begin{figure}
\centering
\begin{tabular}{cc}
	\raisebox{0.2cm}{\rotatebox{90}{Conv}} &
	\includegraphics[width=0.8\linewidth]{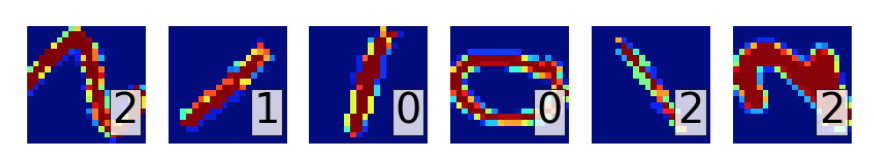} \\
	\raisebox{0.2cm}{\rotatebox{90}{STN}} &
	\includegraphics[width=0.8\linewidth]{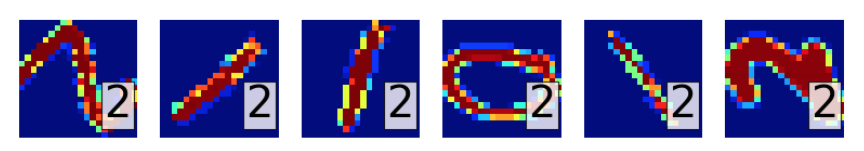} \\
	\raisebox{0.cm}{\rotatebox{90}{TIGraNet}} &
	\includegraphics[width=0.8\linewidth]{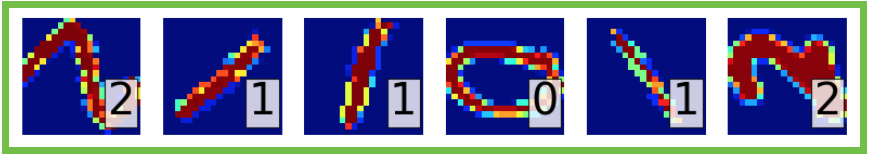} \\
\end{tabular}
	\caption{Illustrative transformation-invariant handwritten digit classification task. Rotated test images, along with their classification label obtained from ConvNets (Conv)~\cite{bb:lecun}, Spatial-Transformer Network (STN)~\cite{bb:STN}, and our method. (best seen in color)}
	\label{fig:adv}
\end{figure}

Deep convolutional networks (ConvNets) have achieved impressive results for various computer vision tasks, such as image classification~\cite{bb:krizhevsky2012imagenetNIPS2012} and segmentation~\cite{bb:seg}. However, they still suffer from the potentially high variability of data in high-dimensional image spaces. In particular, ConvNets that are trained to recognize an object from a given perspective or camera viewpoint, will likely fail when the viewpoint is changed or the image of the object is simply rotated. In order to overcome this issue the most natural step is to extend the training dataset with images of the same objects but seen from different perspectives. This however increases the complexity of data collection and more importantly leads to the growth of the training dataset when the variability of the data is high.

Instead of simply augmenting the training set, which may not always be feasible, one can try to solve the aforementioned problem by making the classification architecture invariant to transformations of the input signal as illustrated in Fig.~\ref{fig:adv}. In that perspective, we propose to represent input images as signals on the grid graph instead of simple matrices of pixel intensities. The benefits of this representation is that graph signals do not carry a strict notion of orientation, while at the same time, signals on a grid graph stay invariant to translation. We exploit these properties to create features that are invariant to isometric transformations and we design new graph-based convolutional and pooling layers, which replace their counterparts used in the classical deep learning settings. This permits preserving the transformation equivariance of each intermediate feature representation under both translation and rotation of the input signals. Specifically, our convolutional layer relies on filters that are polynomials of the graph Laplacian for effective signal representation without computing eigendecompositions of the graph signals. We further introduce a new statistical layer that is placed right before the first fully-connected layer of the network prior to the classification. This layer is specific to our graph signal representation, and in turn permits combining the rotation and translation invariance features along with the power of fully-connected layers that are essential for solving the classification task. We finally design a complete architecture for a deep neural network called TIGraNet, which efficiently combines spectral convolutional, dynamic pooling, statistical and fully-connected layers to process images represented on grid graphs. We train our network in order to learn isometric transformation invariant features. These features are used in sample transformation-invariant image classification tasks, where our solution outperforms the state-of-the-art algorithms for handwritten digit recognition and classification of objects seen from different viewpoints.

In summary, we propose the following contributions in this paper:
\begin{itemize}[topsep=0pt, partopsep=0pt]
	\setlength\itemsep{0pt}
	\setlength{\parskip}{1pt}
	\item We design a new graph-based deep learning framework that learns isometric invariant features;
	\item We propose a new statistical layer that leads to effective transformation-invariant classification of images described by graph-based features;
	\item Through experiments, we show that our method is able to correctly classify rotated or translated images even if such deformations are not present in the training data. 
\end{itemize}

The remainder of the paper is organized as follows. In Section~\ref{s:related_work} we describe the related work. Section~\ref{s:gsp} reviews elements of graph signal processing, which are later used to design graph filters. Our new graph-based architecture is presented in details in Section~\ref{s:overview}. Finally, our experiments and their analysis are presented in Section~\ref{s:exp} and we conclude in Section~\ref{s:conclusion}. 

\section{Related work}
\label{s:related_work}

Most of the recent architectures~\cite{bb:lecun98gradient, bb:krizhevsky2012imagenetNIPS2012} have been very successful in processing natural images, but not necessarily in properly handling geometric transformations in the data. We describe below some of the recent attempts that have been proposed to construct transformation-invariant architectures. We further review quickly the recent works that extend deep learning data represented on graphs or networks. 

\subsection{Transformation-invariant deep learning}

One intuitive way to make the classification architectures more robust to isometric transformations is to augment the training set with transformed data (e.g., \cite{bb:van2012art}), which however, increases both the training set and training time. Alternatively, there have been works that incorporate sort of data augmentation inside the network learning framework. The authors in~\cite{bb:fasel2006rotation} construct deep neural networks that operate in parallel on the original and transformed images simultaneously with weight-shared convolutional filters. Then, the authors in~\cite{bb:dima} extend this multi-column deep neural networks with averaging the output of all the columns to provide the final classification label. A different approach was proposed in~\cite{bb:STN}, where the authors introduce a new spatial transformer layer that deforms images according to a predefined transformation class. Then, the work in \cite{bb:marcos2016learning} suggests using rotated filter banks and a special max pooling operation to combine their outcomes and improve invariance to transformations. The authors in \cite{bb:cohen2016group} propose a generalization of the ConvNets and introduce equivariance to $90\degree$ rotations and flips by exploiting Group Theory arguments in their architecture and introducing a novel G-convolutional layer. Finally, the authors in \cite{bb:dieleman2015rotation} exploit rotation symmetry in the Convolutional Network for the specific problem of galaxy morphology prediction. This work has been extended in~\cite{bb:dieleman-cyclic-2016} which introduces an additional layer that makes the network to be partially invariant to rotations. All the above methods, however, still need to be trained on a large dataset of randomly rotated images in order to be rotation invariant and achieve effective performance.

Contrary to the previous methods, we propose to directly learn feature representations that are invariant to isometric data transformations. With such features, our architecture preserves all the advantages of deep networks, but additionally provides invariance to isometric geometric transformations. The methods in~\cite{bb:oyallon2015deep, bb:bruna2013invariant, bb:harm} are the closest in spirit to ours. In order to be invariant to local transformation, the works in~\cite{bb:oyallon2015deep, bb:bruna2013invariant} propose to replace the classical convolutional layers with wavelets, which are stable to some deformations.
The latter achieves high performance on texture classification task, however it does not improve the performance of supervised ConvNets on natural images, due to the fact that the final feature representations are too rigid and unable to adapt to a specific task. Finally, a very recent work~\cite{bb:harm} proposes a so called Harmonic Network, which uses specifically designed complex valued filters to make feature representations equivariant to rotations. This method, however, still requires the training dataset to contain examples of rotated images to achieve its full potential. On the other hand, we propose building features that are inherently invariant to isometric transformations, which allows us to train more compact networks and achieve state-of-the-art results.

\subsection{Deep learning and graph signal processing}

While there has been a lot of research efforts related to the application of deep learning methods to traditional data like 1-D speech signals or 2-D images, it is only recently that researchers have started to consider the analysis of network or graph data with such architectures \cite{bb:kipf2016semi, bb:henaff2015deep, bb:nips_fingerprint, bb:Structural-RNN}.
The work in \cite{bb:bruna-iclr-14} has been among the pioneering efforts in trying to bridge the gap between graph-based learning and deep learning methods. The authors calculate the projection of graph signals onto the space defined by the eigenvectors of the Laplacian matrix of the input graph, which itself describes the geometry of the data. It however requires an expensive calculation of the graph eigendecomposition, which can be a strong limitation for large graphs, as it requires $O(N^3)$ operations with $N$ being the number of nodes in the graph. 
The authors in \cite{bb:Mikhael} later propose an alternative to analyse network data, which is built on a vertex domain feature representation and on fast spectral convolutional filters. Both methods directly integrate the graph features into a fully-connected layer similarly to classical ConvNets, which is however not directly amenable to transformation-invariant image classification. 

Some recent works further apply deep networks to particular graph data analysis tasks. For example, the authors in~\cite{bb:bronsteingeodesicconv} generalize the ConvNets paradigm to the extraction of feature descriptors for 3D shapes that are defined on different graphs. The work in \cite{bb:nips_fingerprint} further applies deep architectures to train descriptors of chemical molecules, which can be used to predict properties of novel molecules. In~\cite{bb:deepwalk}, the  authors introduce deep networks to analyze web-scale graphs using random walks, which can be used for social network classification tasks. The above algorithms are however specifically developed for a particular task, therefore their generalization to other problems is often difficult.

To the best of our knowledge, the current approaches to deep learning on graphs do not provide transformation-invariance in image classification. At the same time, the methods that specifically target transformation invariance in image datasets mostly rely on data augmentation, which largely remains an art. We propose to bridge this gap and present a novel method that uses the power of graph signal processing to add translation and rotation invariance to the image feature representation learned by deep networks.

\section{Graph signal processing elements}
\label{s:gsp}

We now briefly review some elements of graph signal processing that are important in the construction of our novel framework. We represent an input image as a signal $y(v_n)$ on the nodes $\{v_n\}$ of the grid graph $G$. In more details, $G=\{{\mathcal{V},\mathcal{E}}, A\}$ is an undirected, weighted and connected graph, where $\mathcal{V}$ is a set of $N$ vertices (i.e., the image pixels), $\mathcal{E}$ is a set of edges and $A$ is a weighted adjacency matrix. An edge $e(v_i,v_j)$ that connects two nodes $v_i$ and $v_j$ is associated with the weight $a_{ij}=a_{ji}$, which is usually chosen to capture the distance between both vertices. The edge weight is set to zero for pairs of nodes that are not connected, and all the edge weights together build the adjacency matrix $A$. Every vertex $v_n$ of $G$ carries the luminance value of the corresponding image pixel. Altogether, the valued vertices define a graph signal $y(v_n): \mathcal{V} \to \mathbb{R}$. 

Similarly to regular 1-D or 2-D signals, the graph signals can be efficiently analysed via harmonic analysis and processed in the spectral domain~\cite{bb:shuman2013emerging}. In that respect, we first consider the normalized graph Laplacian operator of the graph $G$, defined as $$\mathcal{L} = I - D^{-1/2} A D^{-1/2},$$ where $D$ is a diagonal degree matrix with elements $d_i = \sum_{n=0, n \neq i}^{N} A_{ni}$. The Laplacian operator is a real symmetric and positive semidefinite matrix, which has a set of orthonormal eigenvectors and corresponding eigenvalues. Let $\chi=[\chi_0, \chi_1, \dots, \chi_{N-1}]$ denote these eigenvectors and $\{0=\lambda_0 \leq \lambda_1\leq \dots \leq \lambda_{N-1} \}$ denote the corresponding eigenvalues with $\lambda_{N-1} = \lambda_{\mathrm{max}} = 2$ for the normalized Laplacian $\mathcal{L}$. The eigenvectors form a Fourier basis and the eigenvalues carry a notion of frequencies as in the classical Fourier analysis. The Graph Fourier Transform $\hat{y}(\lambda_i)$ at frequency $\lambda_i$ for signal $y$ and respectively the inverse graph Fourier transform for the vertex $v_n \in \mathcal{V}$ are thus defined as:
\begin{equation}
\hat{y}(\lambda_i) = \sum_{n=1}^{N} y(v_n) \chi_i^*(v_n),
\end{equation}
and 
\begin{equation}
y(v_n) = \sum_{i=0}^{N-1} \hat{y}(\lambda_i) \chi_i(v_n).
\end{equation}
\noindent

Equipped with the above notion of Graph Fourier Transform, we can denote the generalized convolution of two graph signals $y_1$ and $y_2$ with help of the graph Laplacian eigenvectors as
\begin{equation}
(y_1 * y_2)(v_n) = \sum_{i=0}^{N-1} \hat{y_1}(\lambda_i) \hat{y_2}(\lambda_i) \chi_i (v_n).
\label{eq:conv}
\end{equation}
By comparing the previous relations, we can see that the convolution in the vertex domain is equivalent to the multiplication in the graph spectral domain. Graph spectral filtering can further be defined as
\begin{equation}
\hat{y}_f(\lambda_i) = \hat{y}(\lambda_i) \hat{h}(\lambda_i), 
\end{equation}
where $\hat{h}(\lambda_i)$ is the spectral representation of the graph filter $h(v_n)$ and $\hat{y}_f(\lambda_i)$ is the Graph Fourier Transform of the filtered signal $y_f$. In a matrix form, the graph filter can be denoted by $H \in \mathbb{R^{N \times N}}: H=\chi \hat{H} \chi^T$, where $\hat{H}$ is a diagonal matrix constructed on the spectral representation of the graph filter:
\begin{equation}
\hat{H}=\mathrm{diag} (\hat{h}(\lambda_0), \dots, \hat{h}(\lambda_{N-1})).
\label{eq:hatH}
\end{equation}
The graph filtering process becomes $y_f = H y $, with the vectors $y$ and $y_f$ being the graph signal and its filtered version in the vertex domain. Finally, we can define the generalized translation operator $T_{v_n}$ for a graph signal $y$ as the convolution of $y$ with a delta function $\delta_{v_n}$ centered at vertex $v_n$ \cite{bb:thanou2014learning}:
\begin{equation}
\begin{array}{rl}
T_{v_n} y & = \sqrt{N}(y * \delta_{v_n}) \\
& =\sqrt{N}\sum_{i=0}^{N-1}\hat{y}(\lambda_i)\chi_i^*(v_n)\chi_i .
\end{array}
\label{eq:transl}
\end{equation}
More details about the above graph signal processing operators can be found in \cite{bb:shuman2013emerging}.

\section{Graph-based convolutional network}
\label{s:overview}

\begin{figure*}[tb!]
	\includegraphics[width=1\linewidth]{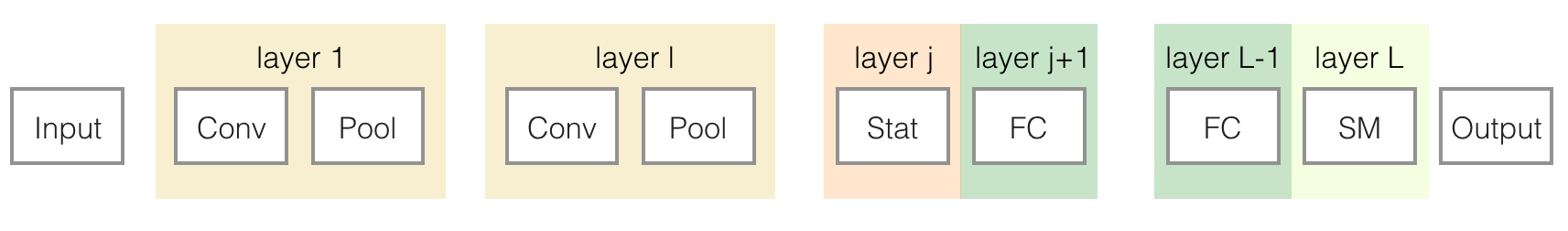}
	\vspace{-1.0cm}
	\caption{{\bf TIGraNet} architecture. The network is composed of an alternation of spectral convolution layers $\mathcal{F}^{l}$  and dynamic pooling layers $\mathcal{P}^{l}$, followed by a statistical layer $\mathcal{H}$, multiple fully-connected layers (FC) and a softmax operator (SM). The input of the network is an image that is represented as a signal $y_0$ on the grid-graph with Laplacian matrix $\mathcal{L}$. The output of the system is a label that corresponds to the most likely class for the input sample.}
	\label{fig:architecture}
\end{figure*}

We now present the overview of our new architecture, which is illustrated in Fig.~\ref{fig:architecture}. The input to our system can be characterized by a normalized Laplacian matrix $\mathcal{L}$ computed on the grid graph $G$ and the signal $y_0 = (y_0(v_1), \dots, y_0(v_N))$, where $y_0(v_j)$ is the intensity of the pixel $j$ in the input image and $N$ is the number of pixels in the images. Our network eventually returns a class label for each input signal.

In more details, our deep learning architecture consists of an alternation of spectral convolution layers $\mathcal{F}^{l}$ and dynamic pooling layers $\mathcal{P}^{l}$. They are followed by a statistical layer $\mathcal{H}$ and a sequence of fully-connected layers (FC) that precedes a softmax operator (SM) that produces a categorical distribution over labels to classify the input data. Both the spectral convolution and the dynamic pooling layers contain $K_l$ operators denoted by $\mathcal{F}_i^{l}$ and $\mathcal{P}_i^{l}, i=1,\dots,K_l$, respectively.  Each convolutional layer $\mathcal{F}_i^{l}$ is specifically designed to compute transformation-invariant features on grid graphs. The dynamic pooling layer follows the same principles as the classical ConvNet's max-pooling operation but preserves the graph structure in the signal representation. Finally, the statistical layer $\mathcal{H}$ is a new layer designed specifically to achieve invariance to isometric transformations on grid graphs. It does not have any correspondent in the classical ConvNets architectures. We discuss more thoroughly each of these layers in the remainder of this section.

\subsection{Spectral convolutional layer}
\label{s:conv}

\begin{figure}[t!]
	\includegraphics[width=1\linewidth]{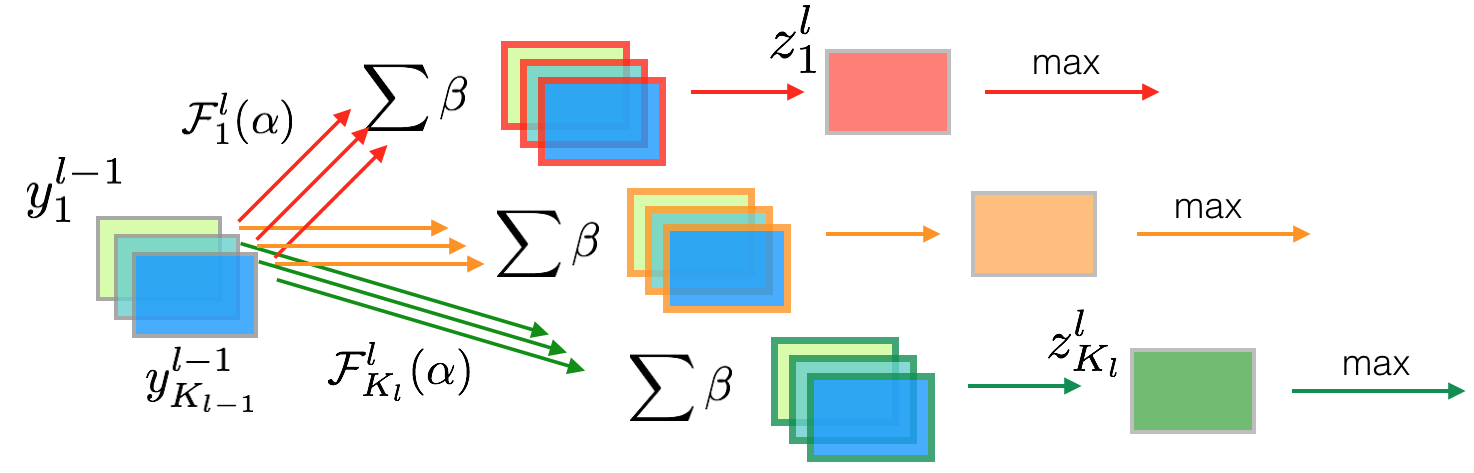}
	\caption{Spectral convolutional layer $\mathcal{F}^{l}$ in {\bf TIGraNet}. The outputs of the previous layer $l-1$ are fed to a set of filter operators $\mathcal{F}_i^{l}$. The outputs of $\mathcal{F}_i^{l}$ are then linearly combined to get the filter maps $z_i^{l}$ that are further passed to the dynamic pooling layer.}
	\label{fig:conv}
\end{figure}

Similarly to the convolutional layers in classical architectures, the spectral convolutional layer $l$ in our network consists of $K_l$ convolutional filters $\mathcal{F}_i^{l}$, as illustrated in Fig.~\ref{fig:conv}. However, each filter $i$ operates in the graph spectral domain. In order to avoid computing the graph eigen-decomposition that is required to perform filtering through Eq. (\ref{eq:conv}), we choose to design our graph filters as smooth polynomial filters of order $M$~\cite{bb:thanou2014learning}, which can be written as
\begin{equation}
\hat{h}(\lambda_l) = \sum_{m=0}^M \alpha_m \lambda_l^m .
\end{equation}
Following the notation of Eq.~(\ref{eq:hatH}), each filter operator in the spectral convolutional layer $l$ can be written as
\begin{equation}
\mathcal{F}_i^{l} = \sum_{m=0}^M \alpha_{i,m}^{l} \mathcal{L}^m ,
\label{eq:pol_filt_1}
\end{equation}
\noindent
where $\mathcal{L}^m$ denotes the Laplacian matrix of power $m$. The polynomial coefficients $\{\alpha_{i,m}^{l}\}$ have to be learned during the training of the network, for each spectral convolutional layer $l$. Each column of this $N \times N$ operator corresponds to an instance of the graph filter centered at a different vertex of the graph~\cite{bb:thanou2014learning}. The support of each graph filter is directly controlled by the degree $M$ of the polynomial kernel, as the filter takes values only on vertices that are less than M-hop away from the filter center. Larger values of $M$ require more parameters but allow training more complex filters. Therefore, $M$ can be seen as a counterpart of the filter's size parameter in the classical ConvNets.

The filtering operation then simply consists in multiplying the graph signal by the transpose of the operator defined in Eq.~(\ref{eq:pol_filt_1}), namely 
\begin{equation}
\tilde{y}_{i,k}^{l} =\left[\mathcal{F}_i^{l}|_{\mathcal{N}_i^{l-1}}\right]^T y_k^{l} , 
\label{eq:pol_filt_2}
\end{equation}
\noindent
where $y_k^{l}$ and $\tilde{y}_{i,k}^{l}$ are the graph signals at the input and respectively the output of the $l^{th}$ spectral convolutional layer (see Fig.~\ref{fig:conv}). In particular, $y_k^{(1)} = y_0$ is the input image for the first level filter, while at the next levels of the network $y_k^{l}$ is rather one of the feature maps output by the lower layers. We finally use the notation $A |_{\mathcal{N}_i^{l}}$ to represent an operator that preserves the columns of the matrix $A$, which have an index in the set ${\mathcal{N}_i^{l}}$, and set all the other columns to zero. This operator permits computing the filtering operations only on specific vertices of the graphs. It is important to note that the spectral graph convolutional filter permits equivariance to isometric transformations, which is a key property for designing a classifier that is invariant to rotation and translation.

Finally, the output of the $l^{th}$ spectral convolutional layer is a set of $K_{l}$ feature maps $z_i^{l}$. Each $i^{th}$ feature map is computed as a linear combination of the outputs of the corresponding polynomial filter as follows: 
\begin{equation}
z_i^{l} =  \sum_{k=1}^{K_{l-1}} \beta_{k}^{l} \ \tilde{y}_{i,k}^{l} , 
\label{eq:lincomb}
\end{equation}
\noindent
where the set of signals $\tilde{y}_{i,k}^{l}$ are the outputs of the $i^{th}$ polynomial filter applied on the $K_{l-1}$ input signals of the spectral convolutional layer with Eq. (\ref{eq:pol_filt_2}). The vector of parameters $\{\beta_{k}^{l}\}$, for each spectral convolutional layer $l$ is learned during the training of the network. The operations in the spectral convolutional layer are illustrated in Fig.~\ref{fig:conv}. Lastly, the complexity of spectral filtering can be computed based on the fact that $\mathcal{L}$ and thus the filters are sparse matrices. Then, the complexity is $O(|\mathcal{E}_M | N)$ where $|\mathcal{E}_M|$ is a maximum number of nonzero elements in the columns of $\mathcal{F}_i^{l}$.

\subsection{Dynamic pooling layer}
\label{s:pool}

In classical ConvNets the goal of pooling layers is to summarize the outputs of filters for each operator at the previous convolutional layer. Inspired by \cite{bb:dmaxpool} we introduce a novel layer that we refer to as dynamic pooling layer, which basically consists in preserving only the most important features at each level of the network.

\begin{figure}[!t]
	\centering
	\includegraphics[width=1\linewidth]{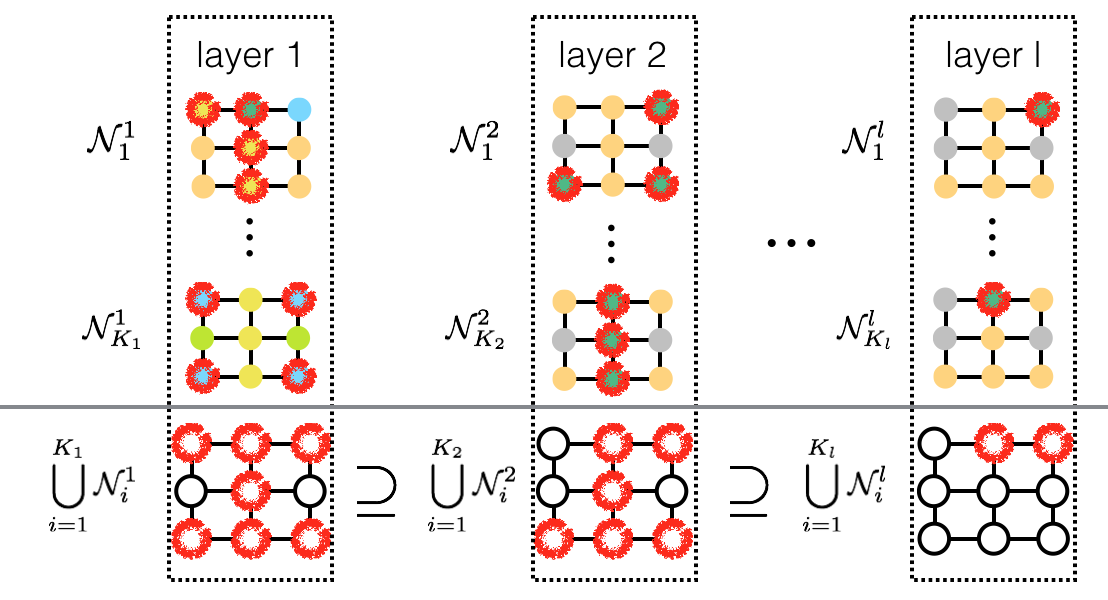} 
	\caption{Pooling process, with succession of dynamic pooling layers with operators $\mathcal{P}_i^{l}$ that each selects the vertices with maximum intensity according to Eq.~(\ref{eq:omega}).
	}
	\label{fig:pool}
\end{figure}

In more details, we perform a dynamic pooling operation, which is essentially driven by the set of graph vertices of interest $\Omega^{l}$. This set is initialised to include all the nodes of graph, i.e., $\Omega^{(1)}= \mathcal{V}$. It is then successively refined along the progression through the multiple layers of the network. More particularly, for each dynamic pooling layer $l$, we select the $J_l$ vertices that are part of $\Omega^{l-1}$ and that have the highest values in $z_i^{l}$. The indexes of these largest valued vertices form a set of nodes $\mathcal{N}_i^{l}$. The union of these sets for the different features maps $z_i^{l}$ form the new set $\Omega^{l}$, i.e., 
\begin{equation}
\Omega^{l} = \bigcup\limits_{i=1}^{K_{l}}\mathcal{N}_i^{l} .
\label{eq:omega}
\end{equation}
The sets $\Omega^{l}$  drives the pooling operations at the next dynamic pooling layer $\mathcal{P}^{l+1}$. We note that, by construction, the different sets $\Omega^{l}$ are embedded, namely we have $\Omega^{l}  \supseteq \Omega^{l+1}, \ \forall l \in [1..L]$. The Algorithm~\ref{alg:pool} summarize our approach, and Fig.~\ref{fig:pool} illustrates the effect of the pooling process through the different network levels.

\begin{algorithm}[h!]
	\begin{algorithmic}[1]
		\STATE {\bf Input:}\quad \ Feature maps $z_i^{l}, i \in [1,K_l]$
		\STATE \qquad\quad\quad Set of nodes of interest $\Omega^{l-1}$
		\STATE \qquad\quad\quad Number of active nodes, $J_l$
		\vspace{5pt}
		\FOR {$i \in [1,K_l]$} 
		\STATE $\overline{\mathcal{N}_i^{l}} = \mathcal{V}$
		\STATE $\mathcal{N}_i^{l} = \emptyset $
		\FOR {$j \in [1,\max{\left(J_l, |\Omega^{l-1} |\right)}]$} 
		\STATE $\nu =  \mathop{\arg\max} \limits_{v \in \left(\Omega^{l-1} \cap \overline{\mathcal{N}_i^{l}} \right)} z_i^{l}(v)$ 
		\STATE $\overline{\mathcal{N}_i^{l}} = \overline{\mathcal{N}_i^{l}} \setminus \{\nu\}$
		\STATE $\mathcal{N}_i^{l} = \mathcal{N}_i^{l} \cup \{\nu\}$
		\ENDFOR
		\ENDFOR
		\STATE $\Omega^{l}=\bigcup\limits_{i=1}^{K_{l}}\mathcal{N}_i^{l}$
	\end{algorithmic}
	\caption{Dynamic pooling layer at layer $l$.}
	\label{alg:pool}
\end{algorithm}

The sets $\mathcal{N}_i^{l}$ are used to control the filtering process at the next layer. The spectral convolutional filters $\mathcal{F}_i^{l+1}$ compute the output of filters centred on the nodes in $\mathcal{N}_i^{l}$ that are selected by the dynamic pooling layer, and not necessarily for all the nodes in the graph. The filtering operation is given by Eq.~(\ref{eq:pol_filt_2}).


Finally, we note that one of the major differences with the classical max-pooling operator is that our dynamic pooling layer is not limited to a small neighbourhood around each node. Instead, it considers the set of nodes of interest $\Omega_l$ which is selected over all graph's nodes. The dynamic pooling operator $\mathcal{P}^{l}$ is thus equivariant to the isometric transformations $R$, similarly to the spectral convolutional layers, which is a key property in building a transformation-invariant classification architecture. The complexity of  $\mathcal{P}^l$ is comparable with the classical pooling operator as the task of $\mathcal{P}^l$ is equivalent to finding $J_l$ highest statistics. Using the selection algorithm~\cite{bb:Knuth98} we can reach the average computational complexity of $O(N)$.

\subsection{Upper layers}
\label{s:hist}

After the series of alternating spectral convolutional and dynamic pooling layers, we add output layers that compute the label probability distributions for the input images. Instead of connecting directly a fully-connected layer as in classical ConvNet architectures, we first insert a new statistical layer, whose output is then fed into fully-connected layers (see Fig.~\ref{fig:architecture}).

The main motivation for the statistical layer resides in our objective of designing a transformation-invariant classification architecture. If fully-connected layers are added directly on top of the last dynamic pooling layers, their neurons would have to memorize large amounts of information corresponding to the different positions and rotation of the visual objects. Instead, we propose to insert a new statistical layer, which computes transformation-invariant statistics of the input signal distributions.

In more details, the statistical layer estimates the  distribution of values on the active nodes after the last pooling layer. The inputs of the statistical layer $j$ are denoted as $\tilde{z}_i$, which correspond to the outputs $z_i^{j-1}$ of the last pooling layer $\mathcal{P}^{j-1}$ where the values on non-active nodes (i.e., the nodes in $\overline{\mathcal{N}_i^{l}}$) are set to zero. We then calculate multiscale statistics of these input features maps using Chebyshev polynomials of the graph Laplacian. These polynomials have the advantage of a fast computation due to their iterative construction, and they can be adapted to distributed implementations \cite{bb:shuman2011chebyshev}. In order to construct these polynomials, we first shift the spectrum of the Laplacian $\mathcal{L}$ to the interval $[-1,1]$, which is the original support of Chebyshev polynomials. Equivalently, we set $\tilde{\mathcal{L}} = \mathcal{L} - I $.

As suggested in \cite{bb:Mikhael}, for each input feature map $\tilde{z}_i$ we iteratively construct a set of signals $t_{i,k}$ using graph Chebyshev polynomials of order $k$, with $k \leq K_{max}$, as
\begin{equation}
t_{i,k} = 2 \tilde{\mathcal{L}} t_{i, k-1} - t_{i, k-2},
\end{equation}
\noindent
with $t_{i, 0} = \tilde{z}_i$ and $t_{i, 1}=\tilde{\mathcal{L}} \tilde{z}_i$.
We finally compute a feature vector that gathers the first order statistics of the magnitude of these signals, namely the mean $\mu_{i,k}$ and variance $\sigma^2_{i,k}$ for each signal $| t_{i,k} |$. This forms a feature vector $\phi_{i}$ of $2 K_{max}+2$ elements, i.e., $\phi_{i} = [\mu_{i,0},\sigma^2_{i,0}, \hdots, \mu_{i,K_{max}},\sigma^2_{i,K_{max}} ]$. We choose these particular statistics as they are prone to efficient gradient computation, which is important during back propagation. Furthermore, we note that such feature vectors are inherently invariant to transformation such as translation or rotation.

The feature vectors $\phi_i$'s are eventually sent to a series of fully-connected layers similarly to classical ConvNet architectures. However, since our feature vectors are transformation invariant, the fully-connected layers will also benefit from these properties. This is in opposition to their counterparts in classical ConvNet systems, which need to compute position-dependent parameters. The details about fully-connected layer parameters are given in the Section~\ref{s:exp}. The output of the fully-connected layers is then fed to a softmax layer \cite{bb:bishop:2006softmax}, which finally returns the probability distribution of a given input sample to belong to a given set of classes. 

\subsection{Training}
We use supervised learning and train our network so that it maximizes the log-probability of estimating the correct class of training samples via logistic regression. Overall, we need to compute the values of the parameters in each convolutional and in fully-connected layers. The other layers do not have any parameter to be estimated.  We train the network using a classical back-propagation algorithm and learn the parameters using ADAM stochastic optimization~\cite{bb:adam}.

We provide more details here about the computation that are specific to our new architecture. We refer the reader to \cite{bb:rumelhart1988learning} for more details about the overall training procedure. The back-propagation in the spectral convolutional layer is performed by evaluating the partial derivatives with respect to the parameters $\alpha: \alpha \in \mathbb{R}^{K_{l-1} \times M}$ of the spectral filters, and to the parameters $\beta: \beta \in \mathbb{R}^{K_{l-1}}$ of the feature map construction. The partial derivatives read
\begin{equation}
	\frac{\partial E} {\partial \alpha^l_{i,m}} = \sum_{k=0}^{K_{l-1}} \beta_k^l \left[\mathcal{L}^m |_{\mathcal{N}_i^{l-1}}\right] y_{k}^{l-1} \frac{\partial E}{\partial z^{l}_i}, \\
	\label{eq:der1}
\end{equation}
\begin{equation}
	\frac{\partial E} {\partial \beta_j^l} =  \sum_{m=0}^M \alpha^l_{i,m} \left[\mathcal{L}^m |_{\mathcal{N}_i^{l-1}}\right] y_{j}^{l-1} \frac{\partial E}{\partial z^{l}_i}, \\
	\label{eq:der2}
\end{equation}
\noindent
where $E$ is the negative log-likelihood cost function, $z^{l}_i=y^{l}_i$ is the output feature map of layer $l$, $K_{l-1}$ denotes the number of feature maps at the previous layer of the network, $M$ is the polynomial degree of the convolutional filter and $\mathcal{L}$ is the Laplacian matrix. Then, we further need to compute the partial derivatives with respect to the previous feature maps as follows
\begin{equation}
	\frac{\partial E}{\partial y^{l-1}_j} = \beta_j^l \sum_{m=0}^M \alpha_{i,m}^l \left[\mathcal{L}^m |_{\mathcal{N}_i^{l-1}}\right] \frac{\partial E}{\partial z^{l}_i}. \\
	\label{eq:der_fm}
\end{equation}

Our new dynamic pooling layers, as well as our statistical layer do not have parameters to be trained. Similarly to the max-pooling operator our dynamic pooling layer  permits back-propagation through the active nodes since the gradient is 0 for the non-selected nodes and not zero for the chosen ones.  
Further, the statistical layer back-propagates the gradients as follows:
\begin{equation}
	\frac{\partial{E}} {\partial t_{i, k}}  =\frac{1}{N} \frac{\partial{E}}{\partial \mu_{i,k}},
	\label{eq:der_stat1}
\end{equation}
\begin{equation}
	\frac{\partial{E}} {\partial t_{i, k}}  = \frac{2 (N-1)}{N^2} \sum_{i=1}^N \left( t_{i, k} - \mu_{i,k} \right) \frac{\partial E}{\partial \sigma^2_{i, k}},\\
	\label{eq:der_stat2}
\end{equation}
\noindent
where $\mu_{i,k}, \sigma^2_{i,k}$ are the inputs to the first fully-connected layer and the outputs of the statistical layer. The derivatives $\partial E / \partial \tilde{z}_i$ are then computed as:
\begin{equation}
	\frac{\partial E}{\partial \tilde{z}_i} = \sum_{k=0}^{K_{max}}\frac{\partial E}{\partial t_{i,k}} \frac{\partial t_{i,k}}{\partial \tilde{z}_i},
\end{equation}
\noindent
where $\partial t_{i,k} / \partial \tilde{z}_i$ are simply the derivatives of Chebyshev polynomials~\cite{bb:shuman2011chebyshev} with maximum order $K_{max}$. Please note that we use the non-linear absolute function $|t_{i,k}|$ before statistical layer, therefore, the gradient at $t_{i,k}=0$ is not defined. In practice, however, we set it to $0$, which gives us a nice property of encouraging some  feature map values to be $0$ and favors sparsity.

Finally, the parameters of the fully-connected layers are trained in a classical way, similarly to the training of fully-connected layers in ConvNet architectures \cite{bb:rumelhart1988learning}. 

\section{Experiments}
\label{s:exp}

 In this section we analyze the results and compare our network to the state-of-the-art transformation-invariant classification algorithms. We first describe the experimental settings. We then analyze our architecture and the influence of the different design parameters. Finally we compare our network to the state-of-the-art transformation-invariant classification algorithms.

\begin{table*}[!ht]
	\centering
	\begin{tabularx}{\linewidth}{Xl}
		\toprule
		Method & Architecture \\
		\midrule
		{\bf Experiments on MNIST-012} & \\
		$\quad$ ConvNet~\cite{bb:lecun} & C[3]-P[2]-C[6]-P[2]-FC[50]-FC[30]-FC[10] \\
		$\quad$ STN~\cite{bb:STN} & C[3]-ST[6]-C[6]-ST[6]-FC[50]-FC[30]-FC[10] \\
		$\quad$ TIGraNet & SC[3, 3]-DP[300]-SC[6, 3]-DP[100]-S[10]-FC[50]-FC[30]-FC[10] \\
		\midrule
		{\bf Other experiments} & \\
		$\quad$ ConvNet~\cite{bb:lecun} & C[10]-P[2]-C[20]-P[2]-FC[500]-FC[300]-FC[100] \\
		$\quad$ STN~\cite{bb:STN} & C[10]-ST[6]-C[20]-ST[6]-FC[500]-FC[300]-FC[100] \\
		$\quad$ DeepScat~\cite{bb:oyallon2015deep} & W[2, 5]-PCA[20] \\
		$\quad$ HarmNet~\cite{bb:harm}  & HRC[1, 10]-HCN[10]-HRC[10, 10]-HRC[10, 20]-HCN[20]-HRC[20, 20] \\
		$\quad$ TIGraNet & SC[10, 4]-DP[600]-SC[20, 4]-DP[300]-S[12]-FC[500]-FC[300]-FC[100] \\
		\bottomrule
	\end{tabularx}
	\caption{Architectures used for the experiments on~\cite{bb:ETH80}. We use the following notations to describe the architectures of the networks. C[$X_1$], P[$X_2$], FC[$X_3$] correspond to the convolutional, pooling and fully-connected layers respectively, with $X_1$ being the number of $3 \times 3$ filters, $X_2$ -- the size of the max-pooling area and $X_3$ -- the number of hidden units. ST[$X_4$] denotes the spatial transform layer with $X_4$ affine transformation parameters. W[$O, J$] and PCA[$X_5$] denote the parameters of DeepScat network with wavelet-based filters of order $O$ and maximum scales $J$, with dimension of the affine PCA classifier $X_5$. HRC[$X_6, X_7$] depicts the harmonic cross correlation filter operating on the $X_7$ neighborhood with $X_6$ feature maps. HCN[$X_8$] is the complex nonlinearity layer of HarmNet with $X_8$ parameters. Finally, SC[$K_l$, $M$] is a spectral convolutional layer with $K_l$ filters of degree $M$, DP[$J_l$] is a dynamic pooling that retains $J_l$ most important values. S[$K_{max}$] is a statistical layer with $K_{max}$ the maximum order of Chebyshev polynomials.}
	\label{tab:arch_part2}
\end{table*}

\subsection{Experimental settings}
\label{s:init}

The initialization of the system may have some influence on the actual values of the parameters after training. We have chosen to initialize the parameters $\alpha_{i,m}^l$~(Eq. \ref{eq:pol_filt_1}) of our spectral convolutional filters so that the different filters uniformly cover the full spectral domain. We first create a set of $Z$ overlapping rectangular functions $w(\lambda, a_i, b_i)$

\begin{equation}
	w(\lambda, a_i, b_i) = 
	\begin{cases} 
		1 & \mbox{if } a_i < \lambda < b_i, \\ 
		0 & \mbox{otherwise.}
	\end{cases}
\end{equation}

The non-zero regions for all functions have the same size, and the set of functions covers the full spectrum of the normalized laplacian $\mathcal{L}$, i.e., $[0, 2]$. We finally approximate each of these rectangular functions by a $M$-order polynomial, which produces a set of initial coefficients $\alpha_{i,m}^l$ that are used to define the initial version of the spectral filter $\mathcal{F}_i^{l}$. Then, the initial values of the parameters $\beta$ in the spectral convolutional layer are distributed uniformly in $[0, 1]$ and those of the parameters in the fully-connected layers are selected uniformly in $[-1, 1]$. 

We run experiments with different numbers of layers and parameters. For each architecture, the network is trained using back-propagation with Adam~\cite{bb:adam} optimization. The exact formulas of the partial derivatives are provided in the supplementary material. 

Our architecture has been trained and tested on different datasets, namely:
\begin{itemize}[topsep=0pt, partopsep=0pt]
	\setlength\itemsep{0pt}
	\setlength{\parskip}{1pt}
	\item \textbf{MNIST-012.} 
This is a small subset of the MNIST dataset \cite{bb:lecun-mnisthandwrittendigit-2010}. It includes 500 training, 100 validation and 100 test images selected randomly from the MNIST images
	 with labels `0', `1' and `2'. This small dataset permits studying the behavior of our network in detail and to analyze the influence of each of the layers on the performance.
	\item
	\textbf{Rotated and translated MNIST.} 
   To test the invariance to rotation and translation of the objects in an image
    we create MNIST-rot and MNIST-trans datasets respectively.
    Both of these datasets contain 50k training, 3k validation and $\sim$9k test images. We use all MNIST digits \cite{bb:lecun-mnisthandwrittendigit-2010} except `9' as it is rotated version resembles `6'.
    In order to be able to apply transformation to the digits, we resize the MNIST-rot to the size $26 \times 26$ and MNIST-trans to the $34 \times 34$.
    The training and validation data of these datasets contain images of digits without any transformation. However, the testing set of  MNIST-rot contains randomly rotated digits by angles in range $[0\degree, 360\degree]$, while the testing set of MNIST-trans comprises randomly translated MNIST examples up to $\pm 6$ pixels in both vertical and horizontal directions.
	\item
	\textbf{ETH-80.} 
	This dataset~\cite{bb:ETH80} contains images of $80$ objects that belong to $8$ classes. Each object is represented by $41$ images captured from different viewpoints located on a hemisphere. The dataset shows a real life example where isometric transformation invariant features are useful for the object classification. We resize the images to $[50 \times 50]$ and randomly select $2300$ and $300$ of them as the training and validation sets and we use the rest of the images for testing.
\end{itemize}
\noindent
For all these datasets, we define $G$ as a grid graph where each node corresponds to a pixel location and is connected with 8 its nearest neighbors with a weight that is equal to $1$. The pixel luminance values finally define the signal $y$ on the graph $G$ for each image. 

\subsection{TIGraNet Analysis}

We analyze the performance of our new architecture on the MNIST-012 dataset. We first give some examples of feature maps that are produced by our network. We then illustrate the spectral kernels learned by our system, and discuss the influence of dynamic pooling operator.

\begin{figure}
	\includegraphics[width=1\linewidth]{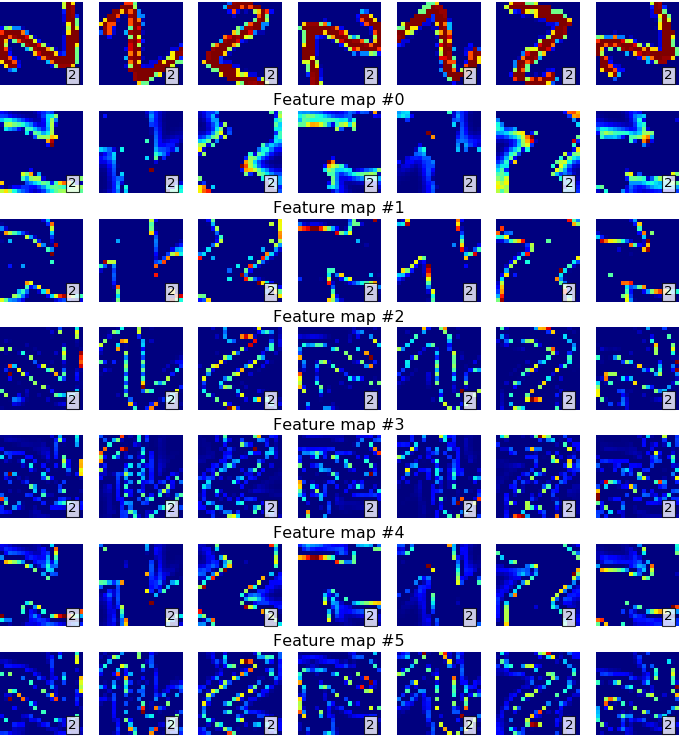} 
	\caption{{\bf Feature maps} from the second spectral convolutional layer for test images that are rotated versions of an image of the digit `2'. The predicted label for each of the images is further shown in the right bottom corner of each image.}
	\label{fig:example_fm}
	\label{fig:VFM}
\end{figure}

We first confirm the transformation invariant properties of our architecture. Even though our classifier is trained on images without any transformations, it is able to correctly classify rotated images in the test set, since our spectral convolutional layer learns filters that are equivariant to isometric transformations. We illustrate this in Fig.~\ref{fig:VFM}, which depicts several examples of feature maps $y_i^2$ from the second spectral convolutional layer for randomly rotated input digits in the test set. Each row of Fig.~\ref{fig:VFM} corresponds to images of a different digit, and we see that the corresponding feature maps are very close to each other (up to the image rotation) even when the rotation angle is quite large. This confirms that our architecture is able to learn features that are preserved with rotation, even if the training has been performed on non-transformed images. Despite important similarities in feature maps of rotated digits, one may however observe some slightly different values for the intensity. This can be explained by the fact that rotated versions of the input images may differ a bit from the original images due to interpolation artifacts.

\begin{figure}[t!]
	\begin{tabular}{cc}
		\includegraphics[width=0.45\linewidth]{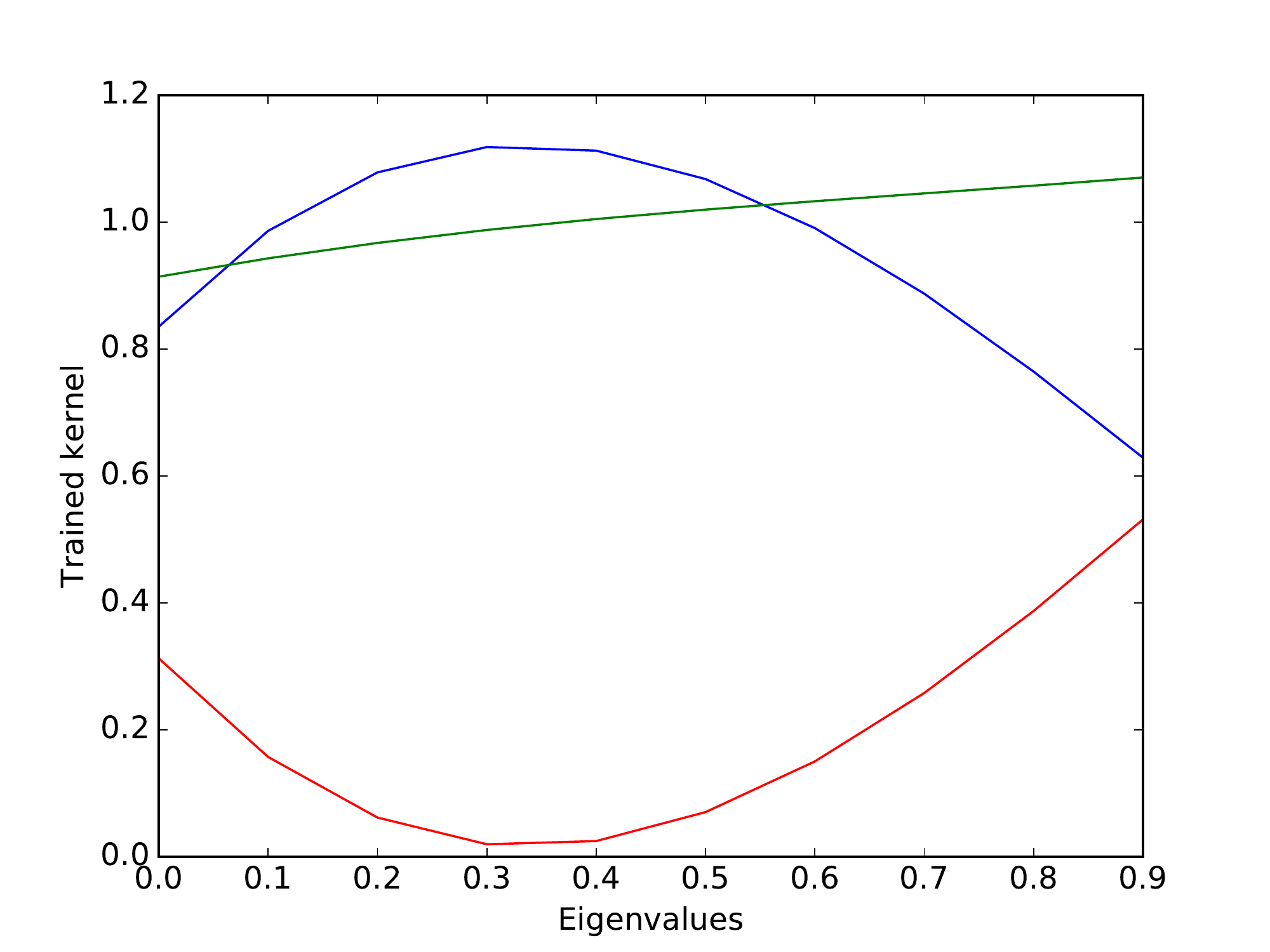} &
		\includegraphics[width=0.45\linewidth]{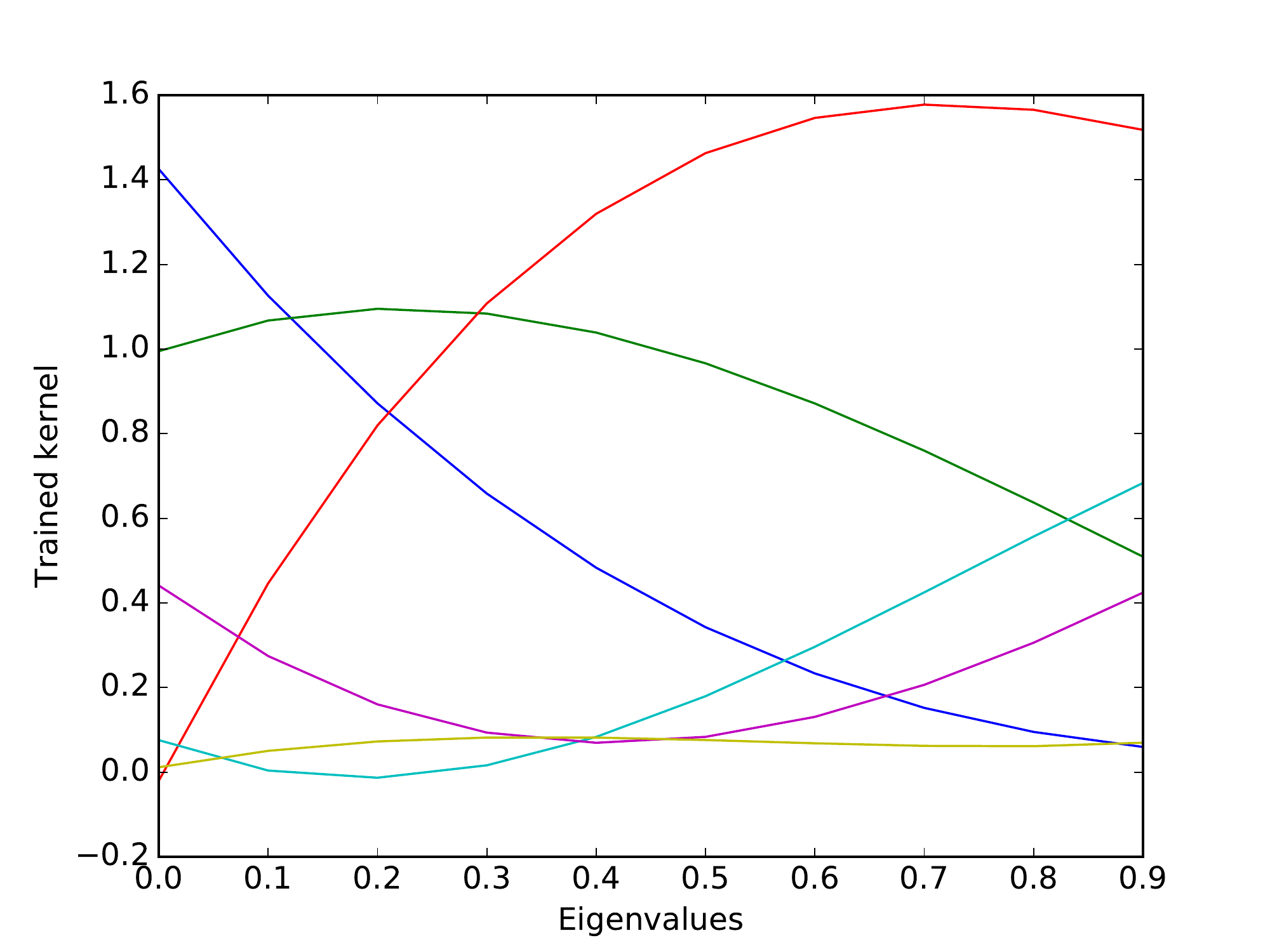}  \\
		a) & b) \\
	\end{tabular}
	\caption{{\bf Sample trained filters} in the spectral domain for (a) first and (b) second convolutional layers. Different colors represent different filters on each of the layers.} 
	\label{fig:filter_ex}
\end{figure}

Fig.~\ref{fig:filter_ex} then shows the spectral representation of the kernels learned for the first two spectral convolutional layers of our network. As expected, the network learns filters that are quite different from each other in the spectral domain but that altogether cover the full spectrum. They permit to efficiently combine information in the different bands of frequency in the spectral representation of the input signal. Generally, the filters in the upper spectral convolutional layers are more diverse and represent more complicated features than those for the lower ones. 

\begin{figure}[t!]
	\includegraphics[width=0.98\linewidth]{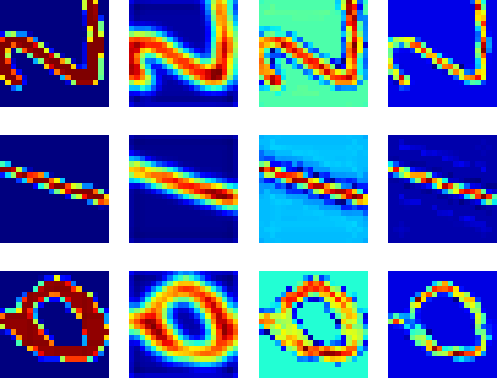}
	\caption{{\bf Feature maps after pooling} Each row shows different digits. The left most column depicts the original images, while the other columns show the features maps after dynamic pooling at the first, second and third layers respectively. The degree of the polynomial filters has been set to $M=3$ for each layer in this experiment.}
	\label{fig:feature_maps_examples}
\end{figure}

Finally, we look at the influence of the new dynamic pooling layers in our architecture. Recall that dynamic pooling is used to reduce the network complexity and to focus on the representative parts of the input signal. Fig.~\ref{fig:feature_maps_examples} depicts the intermediate feature maps of the network for sample test images. We can see that after each pooling operation the signal is getting more and more sparse, while structure of the data that is important for discriminating images in different classes is preserved. That shows that our dynamic pooling operator is able to retain the important information in the feature maps constructed by the spectral convolutional layers.

\subsection{Performance evaluation}

Here, we compare TIGraNet to state-of-the art algorithms for transformation-invariant image classification tasks, i.e., ConvNet~\cite{bb:lecun}, Spatial Transformer Network (STN)~\cite{bb:STN}, Deep Scattering (DeepScat)~\cite{bb:oyallon2015deep} and Harmonic Networks (HarmNet)~\cite{bb:harm}. Briefly, ConvNet is a classical convolutional deep network that is invariant to small image translations. STN compensates for image transformations by learning the affine transformation matrix. Further, DeepScat uses filters based on rich wavelet representation to achieve transformation invariance, however, it does not contain any parameters for the convolutional layers. Finally, HarmNet trains complex valued filters that are equivariant to signal rotations. For the sake of fairness in our comparisons, we use versions of these architectures that have roughly the same number of parameters, which means that each of the approaches learns features with a comparable complexity. For the DeepScat we use the default architecture. Further for the HarmNet we preserve the default network structure, keeping the same number of complex harmonic filters, as the number of spectral convolutional filters that we have in TIGraNet.

We first compare the performance of our algorithm to the ones of ConvNet and STN for the small digit dataset MNIST-012. The specific architectures used in this experiments are given in Table~\ref{tab:arch_part2}.

\begin{table}[t!]
	\centering
	\begin{tabularx}{\linewidth}{ X c c c }
		\toprule
		& \scriptsize{Training set} & \scriptsize{Validation set}  & \scriptsize{Rotated test set} \\
		\midrule
		\multicolumn{4}{l}{\scriptsize{\bf{Training set with data augmentation}}}\\
		\qquad ConvNet &  99  &  94 &  $ 78 \pm 2.1$   \\
		\qquad STN & 100  &  97 & $ 93 \pm 0.97$   \\
		\midrule
		\multicolumn{4}{l}{\scriptsize{\bf{Training set without data augmentation}}}\\
		\qquad ConvNet & 100  &  100 &  $ 55 \pm 5$   \\
		\qquad STN & 100  &  98 & $ 50 \pm 5$   \\
		\qquad  TIGraNet & 98  &  97 & \bf{ 94 $\pm$ 0.42 }   \\
		\bottomrule
	\end{tabularx}
	\caption{Classification accuracy of ConvNet, STN and TIGraNet on MNIST-012. The methods are trained without and with transformed images. We average the performance of all the methods across 10 runs with different transformations of the test data.}
	\label{tab:comparison}
\end{table}
\begin{table}[t!]
	\centering
	\begin{tabularx}{\linewidth}{ X c c }
		\toprule
		& \scriptsize{MNIST-rot} & \scriptsize{MNIST-trans} \\
		\midrule
		\qquad ConvNet & 44.3 & 43.5  \\
		\qquad STN &  44.5 & 67.1 \\
		\qquad TIGraNet &  \bf{83.8} & \bf{79.6} \\
		\bottomrule
	\end{tabularx}
	\caption{Evaluation of the accuracy of the ConvNet, STN and TIGraNet on the MNIST-rot and MNIST-trans datasets. All the methods are trained on sets without transformed images.} 
	\label{tab:all_digit}
\end{table}

The results of this first experiment are presented in Table.~\ref{tab:comparison}. We can see that if we train the methods on the dataset that does not contain rotated images and test on the rotated images of digits, our approach achieves a significant increase in performance (i.e., $86\%$), due to its inherent transformation invariant characteristics. We further run experiments where a simple augmentation of the training set is implemented with randomly rotated each image of digits. This permits increasing the performance of all algorithms, as expected, possibly at the price of more complex training. Still, due to the rotation invariant nature of its features, TIGraNet is still able to achieve higher classification accuracy than all its competitors. 

We then run experiments on the MNIST-rot and MNIST-trans datasets. Note that both of them do not contain any isometric transformation in training and validation sets, but the test set contains transformed images. For all the methods we have used the architectures defined in Table~\ref{tab:arch_part2}. Table~\ref{tab:all_digit} shows that our algorithm significantly outperforms the competitor methods on both datasets due to its transformation invariant features. The other architectures have only limited capabilities with respect to such transformation as rotation. 
 
\begin{figure}
	\centering
	\begin{tabular}{cc}
		\toprule
		\raisebox{2.5cm}{\rotatebox{90}{first layer}} &
		\hspace{-0.2cm} 
		\includegraphics[width=0.9\linewidth]{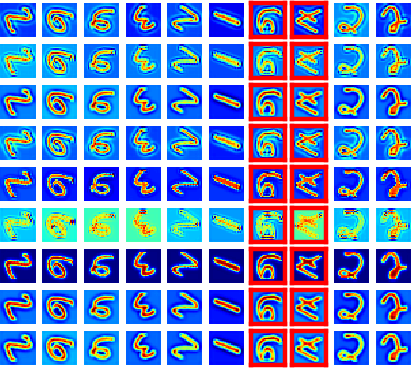} \\
		\midrule
        \raisebox{5.5cm}{\rotatebox{90}{second layer}} &
 		\hspace{-0.2cm} 
		\includegraphics[width=0.9\linewidth]{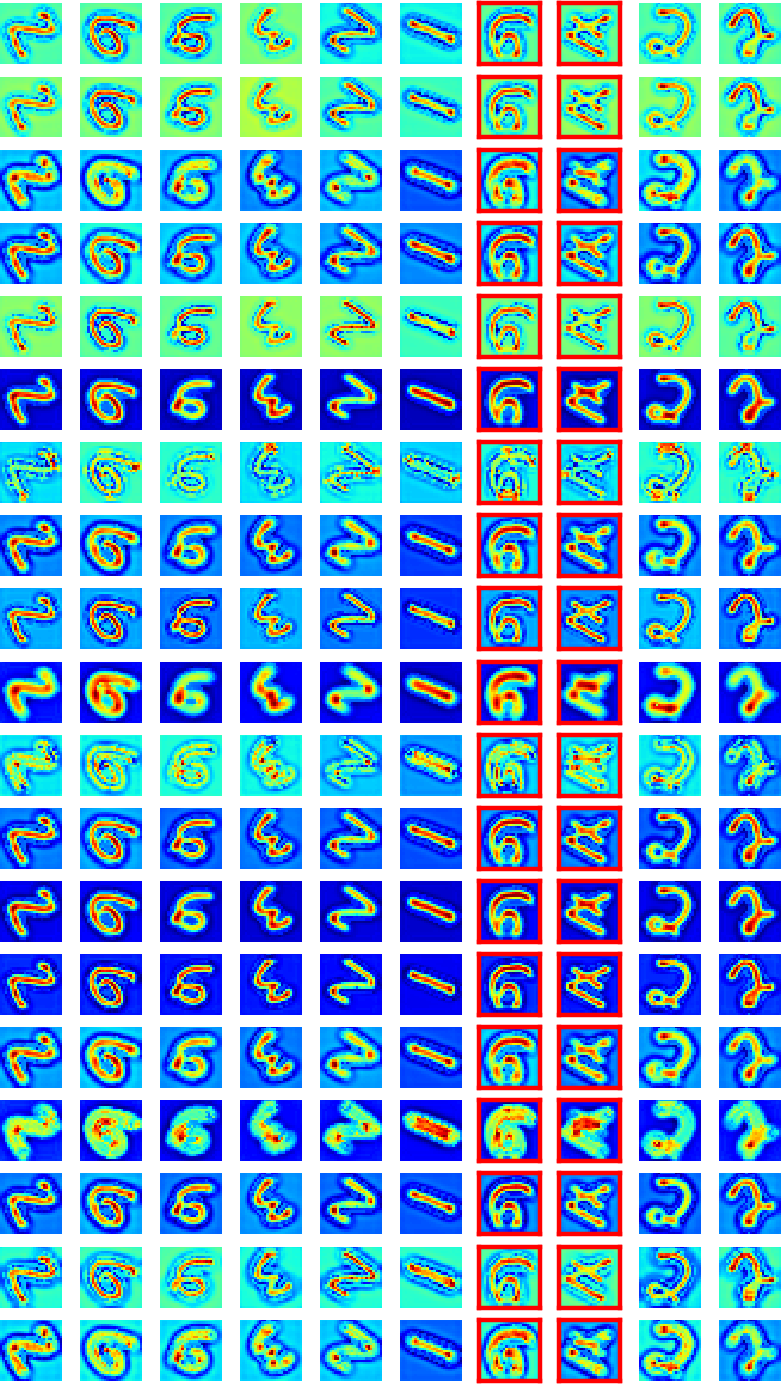} \\
		\bottomrule
	\end{tabular}
	\caption{{\bf Network feature maps visualization.} Each row shows the feature maps of different digits after the first and the second spectral convolutional layers. The misclassified images are marked by red bounding boxes. (best seen in color)
	}
	\label{fig:il_big}
\end{figure}

To further analyze the performance of our network we illustrate several sample feature maps for the different filters of the first two spectral convolutional layers of TIGraNet in Fig.~\ref{fig:il_big}, for the MNIST-rot and MNIST-trans datasets. We can see a few examples of misclassification of our network; for example, the algorithm predicts label `5' for the digit `6'. This mostly happens due to the border artifacts; if the digit is shifted too close to the border due to an isometric transformation, then the neighborhood of some nodes may change. 
This problem can be solved by increasing the image borders or applying filters only to the central pixel locations. 

\begin{table}[!t]
	\centering
	\begin{tabularx}{\linewidth}{ X c }
		\toprule
		& Accuracy (\%) \\
		\midrule
		STN~\cite{bb:STN} &  45.1  \\
		ConvNet~\cite{bb:lecun} & 80.1   \\
		DeepScat~\cite{bb:oyallon2015deep} & 87.3 \\
		HarmNet~\cite{bb:harm} & 94.0   \\
		TIGraNet &  \bf{95.1} \\
		\bottomrule 
	\end{tabularx}
	\caption{Performance evaluation for ConvNet, STN, DeepScat, HarmNet and TIGraNet on classification of images from the ETH-80 dataset.}
	\label{tab:res_eth}
\end{table}

Finally, we evaluate the performance of our algorithm in more realistic settings where the objective is to classify images of objects that are captured from different viewpoints. This task requires having a classifier that is invariant to isometric transformations of the input signal. We therefore run experiments on the ETH-80 dataset and compare the classification performance of TIGraNet to those of ConvNet, STN, DeepScat and HarmNet. The architectures of the different methods are described in Table~\ref{tab:arch_part2}. 

Table~\ref{tab:res_eth} shows the classification results in this experiment. We can see that our approach outperforms the state-of-the-art methods due to its transformation invariant features. The closest performance is achieved by Harmonic Networks, since this architecture also learns equivariant features. It is important to note that the ETH-80 dataset contains less training examples than other publicly available datasets that are commonly used for the training of deep neural networks. This likely results in decrease of accuracy for methods such as~ConvNets and~STN. On the contrary, our method is able to achieve good accuracy even with small amounts of training data, due to its inherent invariance to isometric transformations.

Overall, all the above experiments confirm the benefit of our transformation invariant classification architecture, which learns features that are invariant to transformation by construction. Classification performance improves with these features, such that the algorithm is able to reach sustained performance even if the training set is relatively small, or does not contain similar transformed images as in the test set. These are very important advantages in practice.

\section{Conclusion}
\label{s:conclusion}

In this paper we present a new transformation invariant classification architecture, which combines the power of deep networks and graph signal processing, which allows developing filters that are equivariant to translation and rotation. A novel statistical layer further renders our full network invariant to the isometric transformations. This permits outperforming state-of-the-art algorithms on various illustrative benchmarks. Our new method is able to correctly classify rotated and translated images even if such transformed images do not appear in the training set. This confirms its high potential in practical settings where the training sets are limited but where the data is expected to present high variability. 

\section{Acknowledgment}
The authors thank Dr Dorina Thanou, Damian Foucard and Dr Andreas Loukas for comments that help to improve the paper and we gratefully acknowledge the support of NVIDIA Corporation with the donation of the Tesla K40 GPU used for this research.

\bibliography{egbib_tigra}

\begin{thebibliography}{33}
\providecommand{\natexlab}[1]{#1}
\providecommand{\url}[1]{\texttt{#1}}
\expandafter\ifx\csname urlstyle\endcsname\relax
  \providecommand{\doi}[1]{doi: #1}\else
  \providecommand{\doi}{doi: \begingroup \urlstyle{rm}\Url}\fi

\bibitem[Bishop(2006)]{bb:bishop:2006softmax}
Bishop, C.~M.
\newblock \emph{Pattern Recognition and Machine Learning}.
\newblock Springer, 2006.

\bibitem[Boureau et~al.(2010)Boureau, Ponce, and LeCun]{bb:lecun}
Boureau, Y.~L., Ponce, J., and LeCun, Y.
\newblock {A Theoretical Analysis of Feature Pooling in Visual Recognition}.
\newblock In \emph{International Conference on Machine Learning}, 2010.

\bibitem[Bruna \& Mallat(2013)Bruna and Mallat]{bb:bruna2013invariant}
Bruna, J. and Mallat, S.
\newblock {Invariant scattering convolution networks}.
\newblock \emph{IEEE Transactions on Pattern Analysis and Machine
  Intelligence}, 35\penalty0 (8):\penalty0 1872--1886, 2013.

\bibitem[Bruna et~al.(2014)Bruna, Zaremba, Szlam, and LeCun]{bb:bruna-iclr-14}
Bruna, J., Zaremba, W., Szlam, A., and LeCun, Y.
\newblock {Spectral Networks and Locally Connected Networks on Graphs}.
\newblock In \emph{International Conference for Learning Representations},
  2014.

\bibitem[Cohen \& Welling(2016)Cohen and Welling]{bb:cohen2016group}
Cohen, T.~S. and Welling, M.
\newblock Group equivariant convolutional networks.
\newblock \emph{arXiv preprint}, 2016.

\bibitem[Defferrard et~al.(2016)Defferrard, Bresson, and
  Vandergheynst]{bb:Mikhael}
Defferrard, M., Bresson, X., and Vandergheynst, P.
\newblock {Convolutional Neural Networks on Graphs with Fast Localized Spectral
  Filtering}.
\newblock In \emph{Advances in Neural Information Processing Systems}, pp.\
  3837--3845, 2016.

\bibitem[Dieleman et~al.(2015)Dieleman, Willett, and
  Dambre]{bb:dieleman2015rotation}
Dieleman, S., Willett, K.~W., and Dambre, J.
\newblock Rotation-invariant convolutional neural networks for galaxy
  morphology prediction.
\newblock \emph{Monthly notices of the royal astronomical society},
  450\penalty0 (2):\penalty0 1441--1459, 2015.

\bibitem[Dieleman et~al.(2016)Dieleman, Fauw, and
  Kavukcuoglu]{bb:dieleman-cyclic-2016}
Dieleman, S., Fauw, J.~D., and Kavukcuoglu, K.
\newblock Exploiting cyclic symmetry in convolutional neural networks.
\newblock In \emph{International Conference on Machine Learning}, 2016.

\bibitem[Duvenaud et~al.(2015)Duvenaud, Maclaurin, Iparraguirre, Bombarell,
  Hirzel, Aspuru-Guzik, and Adams]{bb:nips_fingerprint}
Duvenaud, D.~K., Maclaurin, D., Iparraguirre, J., Bombarell, R., Hirzel, T.,
  Aspuru-Guzik, A., and Adams, R.~P.
\newblock {Convolutional networks on graphs for learning molecular
  fingerprints}.
\newblock In \emph{Advances in Neural Information Processing Systems}, pp.\
  2224--2232, 2015.

\bibitem[Dyk \& Meng(2012)Dyk and Meng]{bb:van2012art}
Dyk, D.~A. and Meng, X.-L.
\newblock {The art of data augmentation}.
\newblock \emph{Journal of Computational and Graphical Statistics},
  10:\penalty0 1--111, 2012.

\bibitem[Fasel \& Gatica-Perez(2006)Fasel and
  Gatica-Perez]{bb:fasel2006rotation}
Fasel, B. and Gatica-Perez, D.
\newblock Rotation-invariant neoperceptron.
\newblock In \emph{International Conference on Pattern Recognition}, volume~3,
  pp.\  336--339, 2006.

\bibitem[Henaff et~al.(2015)Henaff, Bruna, and LeCun]{bb:henaff2015deep}
Henaff, M., Bruna, J., and LeCun, Y.
\newblock Deep convolutional networks on graph-structured data.
\newblock \emph{arXiv preprint}, 2015.

\bibitem[Jaderberg et~al.(2015)Jaderberg, Simonyan, Zisserman, and
  Kavukcuoglu]{bb:STN}
Jaderberg, M., Simonyan, K., Zisserman, A., and Kavukcuoglu, K.
\newblock {Spatial Transformer Networks}.
\newblock \emph{arXiv Preprint}, 2015.

\bibitem[Jain et~al.(2015)Jain, Zamir, Savarese, and Saxena]{bb:Structural-RNN}
Jain, A., Zamir, A.~R., Savarese, S., and Saxena, A.
\newblock Structural-rnn: Deep learning on spatio-temporal graphs.
\newblock \emph{arXiv Preprint}, 2015.

\bibitem[Kalchbrenner et~al.(2014)Kalchbrenner, Grefenstette, and
  Blunsom]{bb:dmaxpool}
Kalchbrenner, N., Grefenstette, E., and Blunsom, P.
\newblock {A Convolutional Neural Network for Modelling Sentences}.
\newblock \emph{arXiv Preprint}, 2014.

\bibitem[Kingma \& Ba(2014)Kingma and Ba]{bb:adam}
Kingma, D. and Ba, J.
\newblock {Adam: A method for stochastic optimization}.
\newblock \emph{arXiv Preprint}, 2014.

\bibitem[Kipf \& Welling(2016)Kipf and Welling]{bb:kipf2016semi}
Kipf, T.~N. and Welling, M.
\newblock Semi-supervised classification with graph convolutional networks.
\newblock \emph{arXiv preprint}, 2016.

\bibitem[Knuth(1998)]{bb:Knuth98}
Knuth, D.~E.
\newblock \emph{{The Art of Computer Programming: Sorting and Searching}}.
\newblock Addison Wesley Longman Publishing Co., Inc., 1998.

\bibitem[Krizhevsky et~al.(2012)Krizhevsky, Sutskever, and
  Geoffrey]{bb:krizhevsky2012imagenetNIPS2012}
Krizhevsky, A., Sutskever, I., and Geoffrey, E.~H.
\newblock {ImageNet Classification with Deep Convolutional Neural Networks}.
\newblock In \emph{Advances in Neural Information Processing Systems}, pp.\
  1097--1105, 2012.

\bibitem[Laptev et~al.(2016)Laptev, Savinov, Buhmann, and Pollefeys]{bb:dima}
Laptev, D., Savinov, N., Buhmann, J.M., and Pollefeys, M.
\newblock {TI-Pooling: Transformation-Invariant Pooling for Feature Learning in
  Convolutional Neural Networks}.
\newblock In \emph{Conference on Computer Vision and Pattern Recognition},
  2016.

\bibitem[LeCun \& Cortes(2010)LeCun and
  Cortes]{bb:lecun-mnisthandwrittendigit-2010}
LeCun, Y. and Cortes, C.
\newblock {MNIST} handwritten digit database, 2010.
\newblock URL \url{http://yann.lecun.com/exdb/mnist/}.

\bibitem[LeCun et~al.(2001)LeCun, Bottou, Bengio, and
  Haffner]{bb:lecun98gradient}
LeCun, Y., Bottou, L., Bengio, Y., and Haffner, P.
\newblock {Gradient-Based Learning Applied to Document Recognition}.
\newblock In \emph{Intelligent Signal Processing}, pp.\  306--351, 2001.

\bibitem[Leibe \& Schiele(2003)Leibe and Schiele]{bb:ETH80}
Leibe, B. and Schiele, B.
\newblock Analyzing appearance and contour based methods for object
  categorization.
\newblock In \emph{Conference on Computer Vision and Pattern Recognition},
  volume~2, pp.\  11--409, 2003.

\bibitem[Marcos et~al.(2016)Marcos, Volpi, and Tuia]{bb:marcos2016learning}
Marcos, D., Volpi, M., and Tuia, D.
\newblock Learning rotation invariant convolutional filters for texture
  classification.
\newblock \emph{arXiv preprint}, 2016.

\bibitem[Masci et~al.(2015)Masci, Boscaini, Bronstein, and
  Vandergheynst]{bb:bronsteingeodesicconv}
Masci, J., Boscaini, D., Bronstein, M.~M., and Vandergheynst, P.
\newblock {Geodesic Convolutional Neural Networks on Riemannian Manifolds}.
\newblock In \emph{International Conference on Computer Vision Workshops}, pp.\
   832--840, 2015.

\bibitem[Oyallon \& Mallat(2015)Oyallon and Mallat]{bb:oyallon2015deep}
Oyallon, E. and Mallat, S.
\newblock {Deep roto-translation scattering for object classification}.
\newblock In \emph{Conference on Computer Vision and Pattern Recognition}, pp.\
   2865--2873, 2015.

\bibitem[Perozzi et~al.(2014)Perozzi, Al-Rfou, and Skiena]{bb:deepwalk}
Perozzi, B., Al-Rfou, R., and Skiena, S.
\newblock {Deepwalk: Online learning of social representations}.
\newblock In \emph{International Conference on Knowledge Discovery and Data
  Mining}, pp.\  701--710, 2014.

\bibitem[Ronneberger et~al.(2015)Ronneberger, Fischer, and Brox]{bb:seg}
Ronneberger, O., Fischer, P., and Brox, T.
\newblock U-net: Convolutional networks for biomedical image segmentation.
\newblock In \emph{Medical Image Computing and Computer-Assisted Intervention},
  volume 9351, pp.\  234--241, 2015.

\bibitem[Rumelhart et~al.(1988)Rumelhart, Hinton, and
  Williams]{bb:rumelhart1988learning}
Rumelhart, D.~E., Hinton, G.~E., and Williams, R.~J.
\newblock Learning representations by back-propagating errors.
\newblock \emph{Cognitive modeling}, 5\penalty0 (3):\penalty0 1, 1988.

\bibitem[Shuman et~al.(2011)Shuman, Vandergheynst, and
  Frossard]{bb:shuman2011chebyshev}
Shuman, D.~I., Vandergheynst, P., and Frossard, P.
\newblock {Chebyshev polynomial approximation for distributed signal
  processing}.
\newblock In \emph{IEEE International Conference on Distributed Computing in
  Sensor Systems}, pp.\  1--8, 2011.

\bibitem[Shuman et~al.(2013)Shuman, Narang, Frossard, Ortega, and
  Vandergheynst]{bb:shuman2013emerging}
Shuman, D.~I., Narang, S.~K., Frossard, P., Ortega, A., and Vandergheynst, P.
\newblock {The emerging field of signal processing on graphs: Extending
  high-dimensional data analysis to networks and other irregular domains}.
\newblock \emph{IEEE Signal Processing Magazine}, 30\penalty0 (3):\penalty0
  83--98, 2013.

\bibitem[Thanou et~al.(2014)Thanou, Shuman, and
  Frossard]{bb:thanou2014learning}
Thanou, D., Shuman, D.~I., and Frossard, P.
\newblock {Learning parametric dictionaries for signals on graphs}.
\newblock \emph{IEEE Transactions on Signal Processing}, 62\penalty0
  (15):\penalty0 3849--3862, 2014.

\bibitem[Worrall et~al.(2016)Worrall, Garbin, Turmukhambetov, and
  Brostow]{bb:harm}
Worrall, D.~E., Garbin, S.~J., Turmukhambetov, D., and Brostow, G.~J.
\newblock Harmonic networks: Deep translation and rotation equivariance.
\newblock \emph{arXiv Preprint}, 2016.

\end{thebibliography}
\bibliographystyle{icml2017}

\end{document}